\documentclass[3p, sort&compress]{elsarticle}
\usepackage{hyperref}

\usepackage{graphicx}
\usepackage{amsmath,amssymb} 
\usepackage{color}
\usepackage{amsmath}
\usepackage{amssymb}
\usepackage[ruled]{algorithm2e}
\usepackage[noend]{algpseudocode}
\usepackage{subfigure}
\usepackage{bm}
\usepackage{enumerate}
\usepackage{mdwlist}

\makeatletter
\def\ps@pprintTitle{%
	\let\@oddhead\@empty
	\let\@evenhead\@empty
	\let\@oddfoot\@empty
	\let\@evenfoot\@oddfoot
}
\makeatother

\begin{document}

\begin{frontmatter}
\title{Exemplar Based Deep Discriminative and Shareable Feature Learning for Scene Image Classification}

\author[mymainaddress]{Zhen~Zuo}
\ead{ZZUO1@e.ntu.edu.sg}
\author[mymainaddress,mysecondaryaddress]{Gang~Wang\corref{mycorrespondingauthor}}
\ead{wanggang@ntu.edu.sg}
\author[mymainaddress]{Bing~Shuai}
\ead{BSHUAI001@e.ntu.edu.sg}
\author[mymainaddress]{Lifan~Zhao}
\ead{ZHAO0145@e.ntu.edu.sg}
\author[mythirdaddress]{Qingxiong~Yang} 
\ead{qiyang@cityu.edu.hk}
\cortext[mycorrespondingauthor]{Corresponding author}

\address[mymainaddress]{Nanyang Technological University, Singapore}
\address[mysecondaryaddress]{Advanced Digital Sciences Center, Singapore}
\address[mythirdaddress]{City University of Hong Kong, Hong Kong}

\begin{abstract}
In order to encode the class correlation and class specific information in image representation, we propose a new local feature learning approach named Deep Discriminative and Shareable Feature Learning (DDSFL). DDSFL aims to hierarchically learn feature transformation filter banks to transform raw pixel image patches to features. The learned filter banks are expected to: (1) encode common visual patterns of a flexible number of categories; (2) encode discriminative information; and (3) hierarchically extract patterns at different visual levels. Particularly, in each single layer of DDSFL, shareable filters are jointly learned for classes which share the similar patterns. Discriminative power of the filters is achieved by enforcing the features from the same category to be close, while features from different categories to be far away from each other. Furthermore, we also propose two exemplar selection methods to iteratively select training data for more efficient and effective learning. Based on the experimental results, DDSFL can achieve very promising performance, and it also shows great complementary effect to the state-of-the-art Caffe features.
\end{abstract}


\end{frontmatter}


\section{Introduction}
\label{Introduction}

Extracting informative, robust, and compact data representation (features) has been considered as one of the key factors for good performance in computer vision. Thus, much effort has been paid on developing efficient and effective features, and the existing methods can roughly be classified into two categories: feature engineering and feature learning. In the last decade, numerous feature engineering methods developed hand-crafted features, such as SIFT \cite{lowe2004distinctive}, and HOG \cite{dalal2005histograms}, have ruled the image representation area. However, such methods are labor-intensive and limited by the designer's ingenuity and prior knowledge. In contrast, to expand the capability and ease of image representation, feature learning methods \cite{krizhevsky2012imagenet,le2011ica,zou2012deep,hinton2006,coates2011analysis,sohn2011efficient,zuo2014learning} aim to automatically learn data adaptive image representations from raw pixel image data. However, these methods are generally poor on extracting and organizing the discriminative information from the data. Meanwhile, most of the learning frameworks operate in unsupervised ways without considering the class label information, which is crucial for image classification. To compensate these weaknesses while maintain the advantages of feature learning, we propose to encode shareable information that exists among groups of classes, and discriminative patterns owned by specific classes in feature learning procedure. 

\begin{figure}[!t]
\begin{center}
\includegraphics[width=0.85\linewidth]{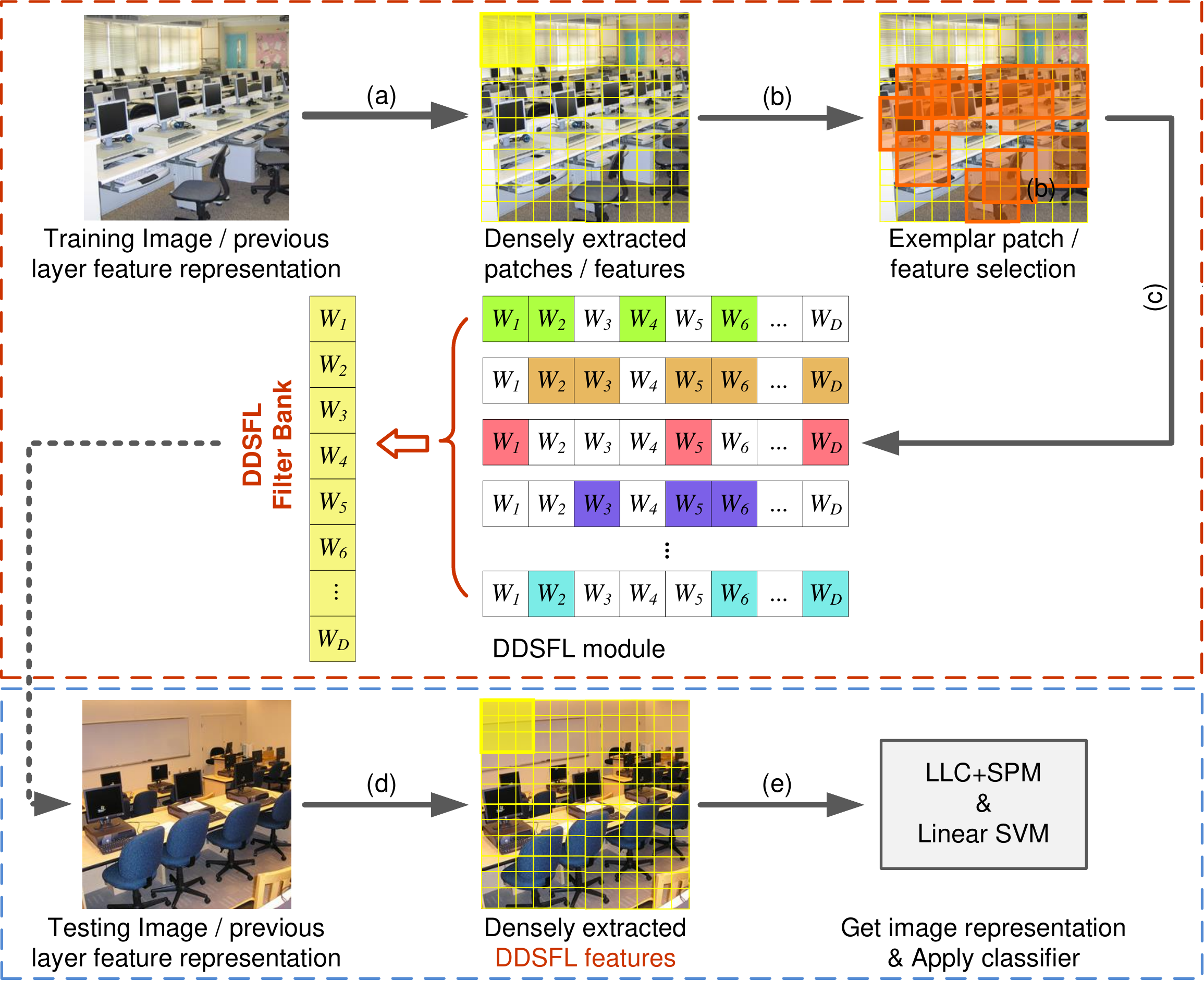}
\vspace{-10pt}
\end{center}
\caption{Illustration of single layer DDSFL. The red dashed box on the top describes the training procedure, while the blue dashed box below indicates the testing process. For the training procedure: (a) Input the original image or feature outputs of the previous layer (if there are any), and densely extract image patches or local features (yellow boxes); (b) Select exemplars (red boxes) for training; (c) Process DDSFL training module and learn filter bank. For the testing procedure: (d) Apply the learned filter bank $W$ to the original input image or previous layer features, and densely extract the DDSFL feature for the current layer; (e) Process LLC \cite{wang2010locality} and SPM \cite{lazebnik2006beyond} afterwards to convert the local features to global image representation, and applying linear SVM to do final classification. In the DDSFL module, $w_1,...,w_D$ represent the filters in the global filter bank $W$. During training step, for each class, we force it to activate a small subset of filters (the activated filters have been highlighted with different colors), and different classes can share the same filters. Finally, the DDSFL feature of a image patch $x_i$ can be represented as ${f_i} = \mathcal{F}\left( {W{x_i}} \right)$. (Best viewed in color)}
\label{DDSFL_flowchart}
\end{figure}

In this paper, we develop a multiple layer feature learning framework called Deep Discriminative and Shareable Feature Learning (DDSFL), which aims to hierarchically learn transformation filter bank to transform pixel values of local image patches to features. As shown in Figure \ref{DDSFL_flowchart}, in each feature learning layer, we aim to learn an over-complete filter bank, which is able to cover the variances of patches from different classes, meanwhile keeping the shareable correlation among similar classes and discriminative power of each category. Intuitively, this goal can be reached by randomly selecting training patches, and learning filter banks for each class independently, then concatenating them together afterwards. However, there are several problems: (1) some of the patterns are shared among some classes, repeatedly learning filters corresponding to the similar patterns are neither memory compact nor computationally efficient, meanwhile the feature dimension will increase linearly with the number of classes, which limit the learning methods to be only applicable to small datasets; (2) discriminative power can hardly be fully exploited, since the class specific characteristics are generally subtle and not obvious without comparing with other classes; (3) in most cases, images are dominated by noisy or meaningless patches, learning filters from randomly sampled image patches will increase the learning cost and depress the performance.

To learn compact and effective filter banks, each category is forced to only activate a subset of the global filters during the learning procedure. Beyond reducing feature dimension, sharing filters can also lead to more robust features. Images belonging to different classes do share some information (e.g. in scene images, both `computer room' and `office' contain `computer' and `desk'). The amount of information sharing depends on the similarity between different categories. Hence, we allow filters to be shareable, meaning that the same filters can be activated by a number of categories. We introduce a binary selection vector to adaptively select which filters to share, and among which categories. 

To improve the discrimination power, we force the features from the same category to be close and the features from different categories to be far away (e.g. patches corresponding to bookshelf in `office' can hardly be found in `computer room'). However, the local patches from the same categories are very diverse. Therefore, we propose to measure the similarity by forcing a patch to be similar to a subgroup of the training samples from the same category instead. Furthermore, not all the local patches from different classes need to be separable. Thus, we relax the discriminative term to allow sharing similar patches across different classes, and focus on separating the less similar patches. 
 
To improve the quality of the filters and efficiency of the learning procedures, we propose two exemplar selection schemes to select effective training data. The proposed methods aim to remove the noisy training patches that commonly exist in many different classes, and select the patches that contain both shareable and discriminative patterns as the training data to learn the filter banks. 

\begin{figure*}
\begin{center}
\includegraphics[width=1\linewidth]{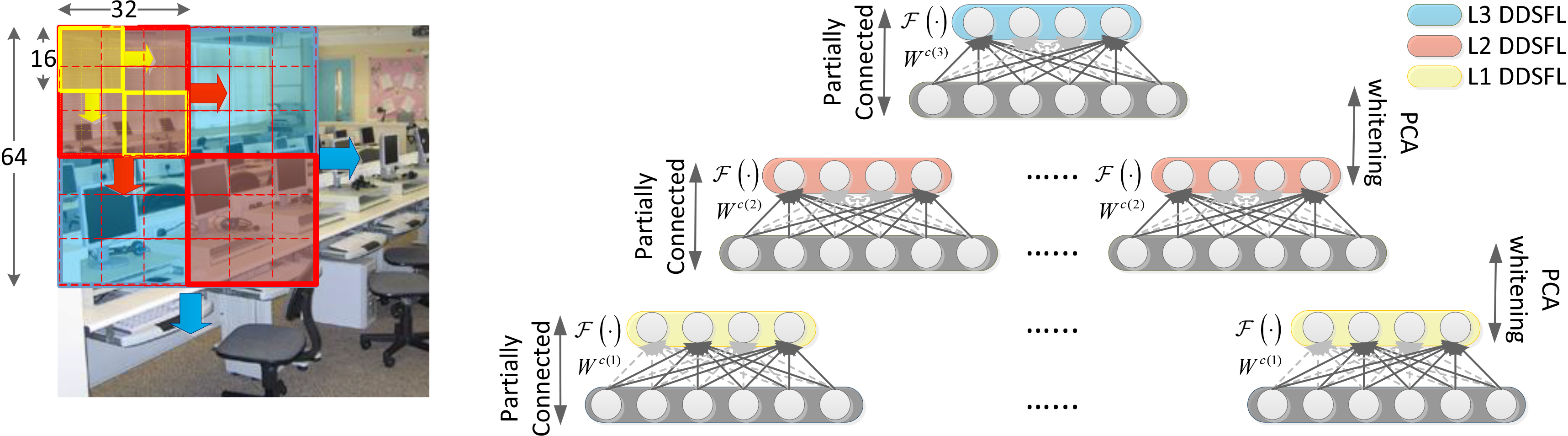}
\end{center}
\vspace{-10pt}
\caption{The hierarchical structure of DDSFL. In the input layer, 16x16 raw image patches are extracted, and be transformed to the first layer features with the first partially connected (in the testing step, fully connected) layer. Afterwards, combining these features correspond to 32x32 image areas, and applying PCA to prepare the reduced dimensional inputs to the second layer. Similarly, the third layer can be processed. Within each partially connected layer, the solid arrows indicate that the corresponding filters are activated for class $c$, while the gray dashed arrows mean filters are not activated. (Best viewed in color)}
\label{DDSFL_structure}
\end{figure*}

Furthermore, supported by lots of previous deep feature learning works, hierarchically extracting increasing visual level features can help to get more abstractive and useful information. Inspired by this idea, we extend the single layer feature learning module to a hierarchical structure. In this paper, we build a three layer learning framework, as shown in Figure \ref{DDSFL_structure}. Specifically, we firstly learn the first layer features from small (16x16) raw pixel value image patches. Then for the higher layers, we convolve the previous layer features within a larger region (32x32 and 64x64 for the second and third layer respectively) as the input to PCA, and use the reduced dimensional data (we set the dimension to 300 for all the layers) as the input to train the current layer filter bank (we set to learn 400 filters for each layer). Finally, we combine the features learned by all the three layers as our DDSFL feature.

The rest of this paper is organized as follows. Section \ref{Related Works} introduces the related works, including feature engineering, feature learning, and discriminative training.  Section \ref{DDSFL Method} describes the details of our DDSFL method by introducing global unsupervised term, shareable term, and discriminative term. Section \ref{Exemplar} proposes two exemplar selection methods including Nearest Neighbor based and SVM based selection. In Section \ref{Experiments}, we test our method on three widely used scene image classification datasets: Scene 15, UIUC Sports, MIT Indoor, and we also test on PASCAL VOC 2012. The experimental results show that our features can outperform most of the existing methods, and it also has significant complementary effect with the state-of-the-art Caffe \cite{donahue2013decaf} features (ConvNets \cite{krizhevsky2012imagenet} pre-trained on ImageNet \cite{deng2010does}). Finally, Section \ref{Conclusion} concludes this paper.

\section{Related Works}
\label{Related Works}

\subsection{Feature Engineering}
In the feature engineering area, hand-crafted features including SIFT \cite{lowe2004distinctive}, HOG \cite{dalal2005histograms}, LBP \cite{ojala2002multiresolution} and GIST \cite{oliva2006building} (global feature) were popular used. Comparing to current existing feature learning methods, they can generally get better local descriptors, and extra information (e.g. discriminative information) can be better expressed by manually inserting prior knowledge. Even though they are very powerful, designing such features is labor intensive, and they can hardly capture any information other than what have been defined by the designers. Therefore, they can hardly extract high visual level patterns. In this paper, we aim to hierarchically learn a data adaptive feature representation.

\subsection{Feature Learning}
In the feature learning field, directly learning features from image pixel values \cite{krizhevsky2012imagenet,le2011ica,zou2012deep,hinton2006,coates2011analysis,sohn2011efficient,le2012deep,le2011learning,zuo2014learning,anran2014learning,wang2014c,yang2015natural,gao2015novel} emerges as a hot research topic in computer vision. These methods are able to learn data adaptive features, and they are usually easy to extend to hierarchical structure, and learn multiple levels of image representations. According to different training settings, there are usually two lines of works: unsupervised/semisupervised feature learning, and supervised feature learning.

Many of the current feature learning frameworks are unsupervised/semisupervised ones, these methods can usually generate numerous additional training data for different objectives, and utilize the intrinsic data similarities to learn transform invariant features. These methods have achieved superior performance on many important computer vision tasks such as image recognition \cite{dosovitskiy2015discriminative,bianco2015curl,wang2014c}, scene classification \cite{yang2015natural,gao2015novel}, and action recognition \cite{le2011learning}. Different from the unsupervised/semisupervised feature learning algorithms, where the class labels are ignored or weakly used, we argue that class specific discriminative information is critical for classification and discriminative patterns can be learned for better image representation. We experimentally show that our DDSFL performs better than unsupervised feature learning on scene datasets by utilizing the shareable and discriminative class correlations. 

In the supervised feature learning line, the ConvNets \cite{krizhevsky2012imagenet} are the most popular deep feature learning structure, they focus on progressively learning multi-levels of visual patterns. ConvNets have been widely utilized to generate global image representations or mid-level discriminative features. They are the state-of-the-art feature learning frameworks on many tasks, such as image recognition \cite{donahue2013decaf,sermanet2013overfeat,girshick2013rich,zuo2015convolutional,dosovitskiy2015discriminative}, scene classification \cite{khan2015discriminative,hayat2015spatial}, scene labeling \cite{shuai2015integrating}, and contour detection \cite{shen2015deepcontour}. In contrast, our DDSFL focuses on encoding the shareable and discriminative correlations among different classes in each layer's feature transformation. Furthermore, different features usually capture different information, fusing complementary features will lead to better performance \cite{wang2014complementary,bianco2015curl}. Although many deep feature learning frameworks are very powerful, they generally focus more on high level image representation, while the low level features (first few layers) are relatively weak. While in this work, our DDSFL framework focus on encoding the class-level discriminative and shareable properties in patch-level local features, which will lead to better image representation for classification tasks. In Section \ref{Experiments}, we will show that our DDSFL learns significant complementary information to the powerful ConvNets features, and combines with which, we can update the current state-of-the-arts on the Scene 15, UIUC Sports, MIT Indoor datasets, and get promising results on PASCAL VOC 2012.

\subsection{Discriminative Training}
There are also some previous works take discriminative information into consideration. For example, \cite{jiang2011learning,mairal2008supervised,yang2011fisher,kong2012dictionary} learn discriminative dictionaries to encode local image features. Another line of works are usually known as mid-level feature searching or exemplar based representation \cite{liharvesting,doersch2013mid,sun2013learning,juneja2013blocks, yao2012action, malisiewicz2011ensemble}. They represent images in terms of weakly-supervise mined discriminative regions of interest (multi-scale patches, usually correspond to middle or high level patterns) or images. Afterwards, images can be represented with the max pooled responses of such mid-level patterns. In contrast, DDSFL focus on discriminatively learning filters, which transform local image patches to features, and allowing sharing local feature transformation filters between different categories. To the best of our knowledge, this hasn't been done before. From another aspect, compared with the previous exemplar based methods, our proposed exemplar selection methods are used to get more informative local patch level training data for the feature learning modules, thus the previous methods (middle or image level exemplar selection) are not suitable in this scenario. Furthermore, in \cite{li2014max}, discriminative information is encoded by maximizing the margin between features from different classes, while making data from the same class to be close. Different from this work, we focus on learning local features for complicated data like scene images, where patches from the same class can be very different, while patches from different classes can be highly similar, thus, max-margin based discriminative learning cannot be directly used here. In \cite{song2012sparselet,song2013discriminatively}, object part filters at the middle level are shared to represent a large number of object categories for object detection. While we only use weakly supervised image level labels, and develop a nearest neighbor based maximum margin method to learn discriminative feature transformation matrix. 

This paper is an extension of our published conference paper \cite{zuo2014learningeccv}. We have proposed a new fast exemplar selection method, provided more details about the previous exemplar selection method, given more details of the algorithm, and extended the previous feature learning framework to a three layer deeper one. We have also provided more complete experiment results: performance on PASCAL VOC 2012, the performance details of using different number of layers, and accuracy of using random initialed filter banks.

\section{Deep Discriminative and Shareable Feature Learning}
\label{DDSFL Method}
In this section, the hierarchical structure of our Deep Discriminative and Shareable Feature Learning (DDSFL) framework will be briefly introduced. For each single layer DDSFL, its three learning components will be introduced in detail, and an alternating optimization strategy will be provided afterwards.

\subsection{DDSFL Hierarchical Structure}
\label{Section3.1}
In this paper, $x_i\in {\mathbb{R}^{D_o}}$ is denoted as the vectorized raw pixel values of an image patch, and $X= \big\{ x_1, x_2, ..., x_N\}$ refer to as the collection of all the $N$ training patches densely extracted from the training images (the patches are weakly labeled by the image level class labels). As shown in Figure \ref{DDSFL_structure}, there are $L=3$ layers in our DDSFL framework. In each layer of DDSFL, we aim to learn a feature transformation filter bank $W^{\left(l\right)} \in {\mathbb{R}^{D \times D_0}},  l=\big\{1,...,L\}$, in which, $l$ is the layer indicator, $D_0$ denotes the dimension of the input data, $D$ is the number of filters, and each row of $W^{\left(l\right)}$ represents a filter. By multiplying $W^{\left(l\right)}$ with an input vector $x_i^{\left(l\right)}$, and applying an activation function $\mathcal{F}\left( \cdot \right)$, we expect to generate feature ${f_i^{\left(l\right)}} = \mathcal{F}\left( {W^{\left(l\right)}{x_i^{\left(l\right)}}} \right)$, which is discriminative and as compact as possible. Afterwards, by convolving the ${f_i^{\left(l\right)}}$in a larger image receptive field, and process PCA, we can get the input ${x_i^{\left(l+1\right)}}$ of the next layer. 

\subsection{Single Layer DDSFL Learning Components}
In each single layer, our DDSFL framework aims to learn features that are able to: 1) preserve the information of the original data; 2) compactly represent patterns shared among similar classes; and 3) effectively encode class specific discriminative patterns. To achieve these goals, three corresponding learning components are proposed: the global unsupervised term, shareable pattern learning term, and discriminative information encoding term.

For the compact consideration, we force each class to activate a subset of filters in ${W^{\left(l\right)}}$, and using the training patches from the similar classes (classes activate the same filters) to train the shared filters. To encode discriminative information, we force the features from the same class to be similar, while features from different classes to be as different as possible. In the following subsections, we will introduce the details of the three learning components, and for the sake of convenience, we ignore the $l$ in all the variables, and focus on the learning procedure of a single layer DDSFL.

\subsubsection{The Global Unsupervised Term}
\label{Section3.1.1}
To ensure the learned filter bank $W$ is able to preserve the information in the original data after processing feature transformation, meanwhile, also be robust to the small variance exist in neighborhood spatial areas, we firstly introduced a global reconstruction term and build an auto-encoder like learning scheme:
\begin{gather}
\begin{split}
& \hspace{0mm} { L_{ u } }=\sum _{ i=1 }^{ N }{ { \mathcal{L} }_{ r }\left( { x }_{ i },W \right) } + \xi \sum _{ i=1 }^{ N }{ { \mathcal{L} }_{ sp }\left( { x }_{ i }, \Omega\left(x_i\right) \right) } +\lambda _{ 1 }\sum _{ i=1 }^{ N }{ { \left\| { f }_{ i } \right\| }_{ 1 } } \\
& \hspace{4mm}{\rm{where}} \hspace{2mm} { \mathcal{L} }_{ r }\left( { x }_{ i },W \right)=\left\| { { x }_{ i }-{ W^{ T } }Wx_{ i } } \right\|_2^2\\
& \hspace{15mm} { \mathcal{L} }_{ sp }\left( { x }_{ i },\Omega\left(x_i\right) \right)=\frac{1}{M}\sum _{ m=1 }^{ M }\left\| { { f }_{ i }- \Omega_m\left( f_i \right)} \right\|_1\\
&\hspace{15mm} {f_i} = \mathcal{F}\left( {W{x_i}} \right) {\rm{,}}\hspace{2mm}\mathcal{F} \left( \cdot \right) = \rm{abs}\left( \cdot \right)
\end{split}
\label{psudo_obj_fun_basic}
\end{gather}
where ${ \mathcal{L}_r }$ is the global reconstruction error with respect to global filter bank $W$ and training data $x_i$. This auto-encoder \cite{hinton2006reducing,le2011ica} style reconstruction cost penalization term can not only prevent $W$ from degeneration, but also allow $W$ to be over-complete. The spatial denoising term $\mathcal{L}_{sp}$ is used to minimize the difference between the training patch and its spatial neighbor patches, and makes the learned features to be robust to small spatial variance\footnotemark[1]. For each training data $x_i$, we randomly sample $M$ patches (extract different size patches, and resize them to have the same size with $x_i$) from the spatial neighborhood image areas, which are denoted as $\Omega\left(x_i\right)$. The term $\left\| { f }_{ i } \right\|_1 $ is used to enforce the sparsity of the learned feature $f_i$. To make the optimization easier while also preserving the quality of the learned features, we use the non-negative activation function in \cite{zou2012deep,le2011learning}, and set $\mathcal{F}\left( \cdot \right) = \rm{abs}\left( \cdot \right)$. Then the sparse term $\left\| { f }_{ i } \right\|_1$ degenerates to the summation of all the elements in vector $f_i$. 

\footnotetext[1]{According to our experiment results, adding the spatial denoising term will not bring much performance gain, but it can help the learning result be more robust to different initialization.}

\subsubsection{Shareable Pattern Learning Term}
\label{Section3.1.2}
Although the above global unsupervised term can already lead to satisfactory results, class information has not been used, which is critical for classification problem. Furthermore, we aim to learn a compact filter bank, rather than a redundant one. Thus, we propose to encode the shareable information exists among similar classes, and utilize the training data from these classes to jointly learn the sharable filters.

To reach this goal, we introduce a sparse binary vector $\alpha^c\in {\mathbb{R}^{D}}$ for each class $c\in\big\{1, ..., C\}$ ($C$ is the number of classes) to indicate the selection status of rows of $W$. For example, if $\alpha_d^{c}=1 ,\hspace{1mm}d = 1,...,D$, then the $d$-th row of $W$ is activated. For each class, only a small number of filters can be activated, while the same filter can be potentially activated by several classes. The cost of our shareable constraint term of class $c$ is formulated as following:
\begin{gather}
\begin{split}
&\hspace{7mm}{ L }_{ \rm{sha} }^{ c }=\sum _{ j=1 }^{ N_c }{ { {\mathcal{L} } }_{ \rm{sha} }^{ c }\left( { x }_j^{ c },W^c \right) } +\lambda _{ 2 }{ \left\| { \alpha }^{ c } \right\| }_{ 0 } \\
&\hspace{0mm}{s.t.}\hspace{2mm}{\alpha _d^{c}} \in \left\{ {0,1} \right\},\hspace{1mm}d = 1,...,D \hspace{5mm}W^c = \rm{diag}\left( {\alpha^c} \right)W\\
& {\rm{where}}\hspace{2mm} {\cal L}_{\rm{sha}}^c\left( { x }_j^{ c }, W^c \right){\rm{ = }}\left\| {{x_j^c} - {{\left( {W^c} \right)}^T} {W^c} {x_j^c}} \right\|_2^2
\end{split}
\label{psudo_obj_fun_sha}
\end{gather}
where $W^c$ is the selected filters of class $c$,  and $N_c$ is the number of training patches from class $c$. Similar to ${ \mathcal{L}_u }$, ${\cal L}_{\rm{sha}}^c$ is the reconstruction cost function with respect to $W^c$ and training patch $x_j^c$ from class $c$. The ${ \left\| { \alpha }^{ c } \right\| }_{ 0 } $ term is used to force the sparsity, consequently, only a small number of rows of $W$ will be activated. Here we apply a greedy method to search for appropriate filters to activate. When $\alpha^c$ is updated in each iteration, the corresponding training data for each filter will also be updated.

\begin{algorithm}[!t]
\DontPrintSemicolon 
\KwIn{\\ $x_i$: Unlabeled training patch\\
$x_j^c$: Image-level labeled training patch from class $c$\\
$D$: Number of filters in the global filter bank\\
$L$: Number of feature transformation layers\\
$C$: Number of the classes\\
$\xi$, $\gamma$, $\eta$, $\lambda_1$, $\lambda_2$: Trade off parameters for controlling weight of spatial denoising term, shareable term, discriminative term, and sparsity}
\KwOut{\\$W^{\left(l\right)}$: Global filter bank (feature transformation matrix) of layer $l$\\}
\BlankLine
1. Initialize $\alpha^c=\mathbf{0}^T$\\
2. Set $W^{\left(l\right)}$ as a random number $D \times D_0$ matrix\\
3. Learn $W^{\left(l\right)}$ with $L_u$ as the initialized $W^{\left(l\right)}$ to the DDSFL\\
4. Select training exemplars for each class (Section \ref{Exemplar})\\
5. Search the positive and negative nearest neighbor sets for each $x_j^c$\\
\For{$l=1$ \rm{to} $L$} {
\While{$W^{\left(l\right)}$ \rm{and} $\alpha^c$ \rm{not converge}} {
\For{$c=1$ \rm{to} $C$} {
6. Fix $W^{\left(l\right)}$ and all the $\alpha^{\bar{c}}$, solve Equation \ref{psudo_obj_fun_2} by updating $\alpha^c$\;
}
7. Fix ${\alpha ^{c}}, {c = 1,...,C} $ and solve Equation \ref{psudo_obj_fun_1} by updating $W^{\left(l\right)}$\\
}
\Return{$W^{\left(l\right)}$}\;}
\caption{Deep Discriminative and Shareable Feature Learning}
\label{algorithm_DSFL}
\end{algorithm}

\subsubsection{Discriminative Information Encoding Term}
\label{Section3.1.3}
To enhance the discriminative power of the features, we further introduce a term based on the assumption that discriminative features should be close to the features from the same category, and be far away from the features from the different categories. However, patches from the same image class are inherently diverse. If we get a triplet $\big\{x_i, x_i^+, x_i^-\}$, in which, $x_i$ denotes the training patch, $x_i^+$ denotes a positive patch randomly select from the same class, and $x_i^-$ represents a negative patch from the other classes, it is highly possible that $x_i$ is similar to $x_i^-$ rather than $x_i^+$. To solve this problem, we adopt the nearest neighbor based `patch-to-class' distance metric \cite{boiman2008defense,mccann2012local,wang2013linear}. For a training patch $x_j^c$, its positive nearest neighbor patch set from the same category is denoted as $\Gamma \left( {x_j^c} \right)$; and its negative nearest neighbor patch set from the categories other than $c$ is denoted as $\bar \Gamma \left( {x_j^c} \right)$. The $k$-th nearest neighbor in the two sets are represented as $\Gamma_k \left( {x_j^c} \right)$ and $\bar \Gamma_k \left( {x_j^c} \right)$ respectively. 

In the class-specific feature space of class $c$ (transformed by $W^c$), the feature representation of the $k$-th positive and negative nearest neighbor patches are denoted as  $\Gamma_k \left( {f_j^c} \right)=\mathcal{F} \left(W^c \Gamma_k\left( {x_j^c} \right)\right)$ and $\bar \Gamma_k \left( {f_j^c} \right)=\mathcal{F} \left(W^c \bar \Gamma_k\left( {x_j^c} \right)\right)$ correspondingly. Then we get triplet $\big\{f_j^c, \Gamma_k \left( {f_j^c} \right), \bar\Gamma_k \left( {f_j^c} \right)\}$, and we aim to minimize the distance between $f_j^c$ and $\Gamma_k \left( {f_j^c} \right)$, while maximize the distance between $f_j^c$ and $\bar\Gamma_k \left( {f_j^c} \right)$. Furthermore, to make the learning procedure more efficient, we should focus on the `hard' training samples (based on the maximum margin theory in learning). Hence, we develop a `hinge-loss' like objective function:
\begin{gather}
\label{discriminative_fun}
\begin{split}
&{L_{{\rm{dis}}}^c} = \sum\limits_{j = 1}^{{N_c}}  \max \left( {\delta  + {\rm{Dis}}\left( {x_j^{\rm{c}},\Gamma \left( {x_j^{\rm{c}}} \right)} \right) - {\rm{Dis}}\left( {x_j^{\rm{c}},\bar \Gamma \left( {x_j^{\rm{c}}} \right)} \right),0} \right)\\
&\hspace{6mm}{\rm{where}} \hspace{1mm}{\rm{Dis}}\left( {x_j^{\rm{c}} , \Gamma \left( {x_j^{\rm{c}}} \right)} \right) = \frac{1}{{{K}}}\sum\limits_{k = 1}^{{K}} {\left\| {f_j^{\rm{c}} - {\Gamma _k}\left( {f_j^{\rm{c}}} \right)} \right\|_2^2} \\
&\hspace{16mm}{\rm{Dis}}\left( {x_j^{\rm{c}},\bar \Gamma \left( {x_j^{\rm{c}}} \right)} \right) = \frac{1}{{{K}}}\sum\limits_{k = 1}^{{K}} {\left\| {f_j^{\rm{c}} - {{\bar \Gamma }_k}\left( {f_j^{\rm{c}}} \right)} \right\|_2^2}\\
\end{split}
\end{gather}
where $K$ is the number of nearest neighbors in the nearest neighbor patch sets, we fixed it as 5, and $\delta$ is the margin, we set it to 1 in our experiments. 

Furthermore, the data in the triplet $\big\{x_j^c, \Gamma_k \left( {x_j^c} \right), \bar\Gamma_k \left( {x_j^c} \right)\}$ depends on the values of $W^c$, which is updating during each iteration. Consequently, when $W^c$ is iteratively being optimized and updated, the nearest neighbors will also be different. Since the nearest neighbor searching is time consuming, we assume the nearest neighbor members $\Gamma_k\left( {x_j^c} \right)$ and $\bar\Gamma_k\left( {x_j^c} \right)$ (the original image patches correspond to the triplet in the feature space) will keep unchanged when $W^c$ does not change too much, and we only update the nearest neighbor members every 50 iterations of $W^c$ updating.

\subsection{DDSFL Objective Function and Optimization}
\label{Section3.4}
Combing the above three learning components, the complete DDSFL objective function can be expressed as:
\begin{gather}
\begin{split}
&\hspace{19mm} \min _{ W, { \alpha }^{ c } }{ L_{ u } } + \gamma \sum\limits_{c = 1}^C {L_{\rm{sha}}^c} {\rm{ + }}\eta \sum\limits_{c = 1}^C {L_{\rm{dis}}^c}  \\
&\hspace{0mm} {\rm{where}} \hspace{2mm} { L_{ u } }=\sum _{ i=1 }^{ N }{ { \mathcal{L} }_{ r }\left( { x }_{ i },W \right) } + \xi \sum _{ i=1 }^{ N }{ { \mathcal{L} }_{ sp }\left( { x }_{ i }, \Omega\left(x_i\right) \right) } +\lambda _{ 1 }\sum _{ i=1 }^{ N }{ { \left\| { f }_{ i } \right\| }_{ 1 } } \\
&\hspace{11mm} { L }_{ \rm{sha} }^{ c }=\sum _{j=1 }^{ N_c }{ { {\mathcal{L} } }_{ \rm{sha} }^{ c }\left( { x }_j^c, W^c \right) } +\lambda _{ 2 }{ \left\| { \alpha }^{ c } \right\| }_{ 0 } \\
&\hspace{11mm} L_{\rm{dis}}^c = \sum\limits_{j = 1}^{{N_c}} {\max } \left( {\delta  + {\rm{Dis}}\left( {x_j^{\rm{c}},\Gamma \left( {x_j^{\rm{c}}} \right)} \right) - {\rm{Dis}}\left( {x_j^{\rm{c}},\bar \Gamma \left( {x_j^{\rm{c}}} \right)} \right),0} \right)\\
&\hspace{2mm}{\rm{s.t.}}\hspace{5mm}{\alpha _{d}^c} \in \left\{ {0,1} \right\},d = 1,...,D \hspace{5mm}W^c = \rm{diag}\left( {\alpha^c} \right)W
\end{split}
\label{full_obj_fun}
\end{gather}

In Equation \ref{full_obj_fun}, we need to simultaneously optimize the global filter transformation matrix $W$ and the class specific filter selection vector $\alpha^c$. This function cannot be jointly optimized with respect to $W$ and $\alpha^c$. However, if one of them is fixed, the objective function turns to be convex. Therefore, we adopt an alternating optimization strategy to iteratively update $W$ and $\alpha^c$.

We firstly fix $\alpha^c$ for each class, and focus on updating $W$:
\begin{gather}
\begin{split}
&{\min _W}\sum\limits_{i = 1}^N {{{\cal L}_r}\left( {{x_i},W} \right)} + \xi \sum _{ i=1 }^{ N }{ { \mathcal{L} }_{ sp }\left( { x }_{ i }, \Omega\left(x_i\right) \right) }  + {\lambda _1}\sum\limits_{i = 1}^N {{{\left\| {{f_i}} \right\|}_1}} \\
& \hspace{6mm}+ \gamma \sum\limits_{c = 1}^C {\sum\limits_{j = 1}^{{N_c}} {{\cal L}_{\rm{sha}}^c\left( {x_j^c, W^c} \right)} } + \eta \sum\limits_{c = 1}^C {L_{\rm{dis}}^c}
\end{split}
\label{psudo_obj_fun_1}
\end{gather}

Because of the non-negative constraint applied on $f_i$ (Section \ref{Section3.1.1}), $\left\| { f }_{ i } \right\|_1$ degenerates to summation of different elements in $f_i$. Consequently, Equation \ref{psudo_obj_fun_1} can be easily optimized by unconstrained solvers like L-BFGS.

Next we fix the $W$, and focus on updating $\alpha^c$ class by class:
\begin{equation}
\centering
\label{psudo_obj_fun_2}
\mathop {\min }\limits_{{\alpha ^c}} \sum\limits_{j = 1}^{{N_c}} {{\cal L}_{\rm{sha}}^c\left( {x_j^c,W^c} \right)} + {\lambda _2}{\left\| {{\alpha ^c}} \right\|_0} + \eta {L_{\rm{dis}}^c}
\end{equation}

We update one $\alpha^c$ each time for the $c$-th class, and keep ${\alpha ^{\bar c }}$ ($\bar c\ne c$) unchanged. To get $\alpha^c$, we apply a greedy optimization method. We first set all the elements in $\alpha^c$ as 0, then we search for the single best filter that can minimize Equation \ref{psudo_obj_fun_2}, and activate that filter by setting the corresponding element in $\alpha^c$ to 1. Afterwards, based on the previously activated filters, we search for next filter that can further minimize the cost function. The optimization terminates when the loss ${ \mathcal{L} }_{ \rm{sha} }^c$ is smaller than a threshold, and the renewed $\alpha^c$ will be sent as the input to Equation \ref{psudo_obj_fun_1} again to further optimize $W$.

The learning algorithm and initialization procedure are shown in Algorithm \ref{algorithm_DSFL}. The alternative optimization terminated until the values of both $W$ and $\alpha^c$ converge (takes about 5 rounds).

\section{Exemplar Selection}
\label{Exemplar}
The quality of the learned features not only depends on the learning structure and parameters, but also relies on the quality of input training data. In order to select training patch set that carries both potential shareable and discriminative patterns, while also excludes common noisy patches, a training data selection procedure is required before processing feature learning.

For this purpose, one intuitive solution is applying k-means clustering, and using the cluster centroids as the exemplars \cite{wang2013linear} for training. However, as analyzed in \cite{boiman2008defense}, informative descriptors have low database frequency, and the conventional clustering methods prefer dominant patterns as inlier of clusters. Consequently, the non-informative patches (e.g. sky areas) are more likely to be selected rather than the discriminative ones. Thus, we propose two discriminative exemplar training data selection approaches: Nearest Neighbor (NN) based and Support Vector Machines (SVM) based exemplar selection. 

\subsection{NN Based Exemplar Selection}
Inspired by the image-level exemplar selection method in \cite{yao2012action}, we propose a NN based exemplar selection scheme to remove the patches that are likely to commonly exist in many classes.

In \cite{yao2012action}, the exemplar images are defined as the minimum set of `dominating images' which can cover all the intra-class variations, meanwhile as distinctive as possible to the exemplars from other classes. Different from this scenario, our exemplars are local image patches/features. Furthermore, we aim to keep shareable patterns, while also select class-specific discriminative patterns. To make the processing more applicable, we manipulate the selection procedure by removing the uninformative common patterns existing in most of the classes.

Firstly, we define the `coverage set' of a patch $x_i$. Given $X$ as the original global patch set, which contains patches densely extracted from all the training images. For each patch $x_i \in X$, we search its $M_{nn}$ nearest neighbors from $X$, and define these $M_{nn}$ patches as the `coverage set' of $x_i$. 

Secondly, we define the `reaching number' and `reaching distance' of a patch $x_j^c$. If $x_j^c$ is included in the `coverage set' of a patch $x_l^{\bar c}$ from classes other than $c$, it is `reached' once, and the times of being reached over the patch set $X^{\bar c}$ (from class $\bar c$) is the `reaching number' of $x_j^c$. Meanwhile, we also record the distances between $x_j^c$ and the patches which cover $x_j^c$. The summation of these distances is the `reaching distance' of $x_j^c$. 

Intuitively, the larger the `reaching number' or the smaller the `reaching distance', the more common the $x_j^c$. However, if we only consider `reaching distance', for the common patterns (e.g. sky area patches), there are numerous similar patches with small variation. Consequently, because of the limited size of `coverage set', the chance that these common patches be `reached' by the other similar common patches are not always high, and a common patch might get small `reaching number', even though its similar patterns widely exist. On the other hand, if we only consider `reaching distance', for a target discriminative patch, it might be only covered once, while the corresponding patches are highly similar to it, then its `reaching distance' is unfairly small. Therefore, we make use of both `reaching number' and `reaching distance', and define the `reaching score' of $x_j^c$ to class $\bar c$ as following:
\begin{gather}
\centering
\begin{split}
& {R_S^{\bar{c}}(x_j^c, \bar c)} = {\frac{1}{R_N(\bar c)}\sum\limits_{l = 1}^{R_N(\bar c)} R_d } \\
& \hspace{1.5ex}{\rm{where}} \hspace{3mm} R_d = {{\left\| x_j^{\rm{c}} - x_l^{{\rm{\bar c}}}\right\|}_2}
\end{split}
\label{exemplar_reaching_score_perC}
\end{gather}
where $R_S^{\bar{c}}$ is the class wise reaching score, $R_N(\bar c)$ is `reaching number'  denotes the number of the patches from classes $\bar c$ whose coverage sets contain $x_j^c$, and summation of $R_d$ is the `reaching distance'. 

\begin{algorithm}[!t]
\DontPrintSemicolon 
\KwIn{\\$X$: Global patch set\\
$X_{c}$: Patch set of class $c$\\
$C$: Number of the classes\\
$M_{nn}$: Number of patches in each coverage set}
$\varepsilon_{nn}$: Threshold for selecting discriminative exemplars\\
\KwOut{\\$E_c$: Exemplars of class $c$\\}
\BlankLine
1. Calculate the coverage set of each patch from $X$\\
\For{$c=1$ \rm{to} $C$} {
2. For each patch from $X_c$, calculate its reaching score ${R_S(x_j^c)}$ based on Equation \ref{exemplar_reaching_score} \\
3. Descendingly sort the patches from $X_c$ based on the reaching scores.\\
4. Select the top $\varepsilon_{nn}$ percent ranked patches as the exemplars $E_c$
}
\Return{$E_c$}
\caption{NN based Exemplar Selection}
\label{algorithm_exemplar_NN}
\end{algorithm}

Furthermore, to keep the shareable patterns, we also prefer to penalize the patterns appearing in most of the classes rather than shared among few classes. Thus, we define the final `reaching score' as:
\begin{equation}
\centering
\label{exemplar_reaching_score}
{R_S(x_j^c)} = \frac{1}{{C - 1}}\sum\limits_{\bar c \ne c} {R_S^{\bar{c}}(x_j^c, \bar c)} 
\end{equation}
where $C$ is the number of classes. If $x_j^c$ has low `reaching score', then it means that $x_j^c$ represents a common pattern, and should be removed in the training procedure, otherwise, $x_j^c$ should be kept an exemplar.

For each class, we sort the patches based on their  ${R_S(x_j^c)}$ scores descendingly, and select the top $\varepsilon_{nn}$ percent of them as discriminative exemplars. The selecting procedures are shown in Algorithm \ref{algorithm_exemplar_NN}.

\subsection{SVM Based Exemplar Selection}
Inspired by the mid-level discriminative exemplar searching methods \cite{singh2012unsupervised, doersch2013mid}, we also try to use learning based methods to search for local patch level exemplars. Compare to the NN based selection method, SVM based method is much faster and cost less memory with small performance drop.

The desired exemplars should be both representative and discriminative: similar patches of the exemplars should be able to be found from the same or similar classes, and the exemplars of a certain class should also contain class-specific patterns. To reach these goals, we model the problem as a discriminative clustering problem, and the centers of the effective clusters are considered as the exemplars. In this paper, we use linear SVM to iteratively refine the selection planes for each cluster. In one iteration of each cluster, there are mainly two steps: SVM training and SVM testing (re-clustering). In the SVM training step, the patches from the same cluster are positive samples, and the patches from all the other classes are negative samples, and a linear SVM model is learned based on them. In the SVM testing step, the previous learned SVM model is used to test on all the validation data patches (not visible in the training procedure), the patches which get fired\footnotemark[2] are kept and be used to replace the data in the original cluster, and they will be used to learn the new SVM model in the next iteration. To ensure the exemplars are representative to the class, we do not consider clusters with less than $\varepsilon_{svm}$ ($\varepsilon_{svm}=3$) members. To ensure the purity of the clusters, we constrain that each effective cluster can have at most $M_{svm}$ (we set $M$ as 10) members. 

\begin{algorithm}[!t]
\DontPrintSemicolon 
\KwIn{\\$X^{tr}$: Training patch set\\
$X_{cs}^{tr}$: The $s$-th cluster of the training patch set from class $c$\\
$X_{\bar{c}}^{tr}$: Training patches from classes other than $c$\\
$X^{val}$: Validation patch set\\
$S_c$: Number of initial clusters for each class $c$\\
$C$: Number of the classes\\
$M_{svm}$: Maximum number of patches in each cluster}
\KwOut{\\$E_c$: Exemplars of class $c$\\}
\BlankLine
1. Applying k-means to initialize $S_c$ clusters for each class\\
\While {\rm{not converged}}{
\For{$c=1$ \rm{to} $C$} {
\For{$s=1$ \rm{to} $S_c$} {
2. For each cluster $s$ from class $c$, set the cluster members $X_{cs}^{tr}$ as positive data and randomly sample patches from $X_{\bar{c}}^{tr}$ as negative data, and train a linear SVM\\
3. Test the learned SVM on the validation set $X^{val}$. If the SVM meets any of the non-discriminative condition, remove the cluster.
}}
4. Swap training set $X^{tr}$ and validation set $X^{val}$.}
5. Use the remaining qualified cluster centers as the exemplars $E_c$\\
\Return{$E_c$}
\caption{SVM based Exemplar Selection}
\label{algorithm_exemplar_SVM}
\end{algorithm}

We randomly pick a large subset of local patches from the training images as exemplar candidates, and run k-means clustering for all of the candidate patches of each class to initialize the clusters. Since the SVM heavily prefers the training data during testing, thus we equally separate the candidate patch set into training set and validation set. For the learned SVMs, we test them on the validation set class by class. If either of the following two conditions is true, then the cluster is considered as not discriminative, and should be removed:
\begin{itemize}
\item The SVM fires\footnotemark[2] less than $\varepsilon_{svm}$ (3 in our experiment) patches from the same class (considered as outliers or noisy data); 
\item The SVM fires\footnotemark[2] many patches from more than half of all the classes (considered as common patterns exist everywhere).
\end{itemize}
Afterwards, we swap validation set and training set, and repeat the above procedures until convergence. The iteration usually terminates in about 5 rounds. The overall procedure is shown in Algorithm \ref{algorithm_exemplar_SVM}.
\footnotetext[2]{Classification score higher than -1} 

\subsection{NN Based vs SVM Based exemplar selection}
We propose NN based and SVM based exemplar selection methods. According to the experiment results (Table \ref{Table-interations}), both of them bring significant improvements than DDSFL without exemplar selection procedure. Here we make a brief comparison between these two methods in three aspects: performance, computation cost, and memory cost.

In classification performance aspect, comparing with the result of randomly select training patch, NN based exemplar selection leads to around 1\% to 3\% improvements, while SVM based selection gets around 0.5-1.5\% improvements. Thus, NN based method is a better choice for higher accuracy results.

Next we compare the two methods in computation cost aspect. For the NN based selection, the most time consuming procedure is the step 1 in Algorithm \ref{algorithm_exemplar_NN}: coverage set searching. To speed up the procedure, we use the FLANN \cite{muja2009fast} toolbox, which is a library utilize multiple randomized k-d trees to achieve fast NN approximation for high dimensional data. For the SVM based selection, step 2 and 3 in Algorithm \ref{algorithm_exemplar_SVM} (SVM training and testing) are the most and second most time consuming steps. To speed up the SVM training procedure, we apply stochastic learning of SVM \cite{wang2012learning}. While in the SVM testing step, for each trained linear SVM model, only classification score of the validation data patches need to be calculated. This procedure can be easily parallel computed, and we use GPU to speed up. In our experiments, the SVM based methods cost much less hours, especially when the number of training patches is very large.

At last, the memory costs are compared. In the NN based selection, we need to keep all the training patches $X$ in the memory to apply coverage set searching for each patch. To depress the memory cost, we separate the $X$ into several subgroups, process NN searching for each subgroup, and resort the searching results from all the subgroups afterwards. However, the more subgroups we generate, the longer overall NN searching time will be cost. On the contrary, SVM based selection requires very limited memory, and only an ignorable number of data points need to be loaded into the memory.

Therefore, NN based method can be used to achieve higher accuracy, while SVM based method can be manipulated faster with much less memory cost with less performance gain.

\section{Experiments and Analysis}
\label{Experiments}

\subsection{Datasets and Experiment Settings}

We tested our DDSFL method on three widely used scene image classification datasets: Scene 15 \cite{lazebnik2006beyond}, UIUC Sports \cite{li2007and}, and MIT Indoor \cite{quattoni2009recognizing}. We also tested on the challenging PASCAL VOC 2012 object classification dataset. To make a fair comparison with other types of features, we only utilized gray scale information for all of the images in these datasets.

\begin{itemize}
\item \textbf{Scene 15:} This dataset includes 4,485 images from 15 outdoor and indoor scene categories, each category contains 200 to 400 gray scale images. Based on the standard setting, we use 100 images per category for training, and the rest images for testing.
\item \textbf{UIUC Sports:} This dataset contains 1,579 from 8 sports event classes, each category contains 137 to 250 images. We use 70 images per class as training images, and 60 images per class as testing images. 
\item \textbf{MIT Indoor:} This dataset have 15,620 image in total, and they are from 67 indoor scene categories. Based on the splitting set in \cite{quattoni2009recognizing}, there are around 80 training images and 20 testing images for each category.
\item \textbf{PASCAL VOC 2012:} This challenge aims to recognize objects from a number of visual object classes in realistic scenes. In this dataset, 20 object classes are selected. There are 5,717 training images with 13,609 objects, 5,823 validation images with 13,841 objects. Thus, the total number of labeled images are 11,540, and we did not use any extra training data. In the testing procedure, 10,991 unlabeled testing images are provided, and the classification results can be achieved by uploading the classification scores to the PASCAL VOC evaluation server.
\end{itemize}

Since the resolution of the original images in UIUC Sports, MIT Indoor and PASCAL VOC 2012 are too high for learning local features efficiently, we resize them to have maximum 300 pixels along the longer axis. For Scene 15 and UIUC sports, we randomly split the training and testing datasets for 5 times, and get the results of different methods based on the same splits. The average accuracy numbers over these 5 rounds are reported for comparison.

For all the local features, we densely extracted features from $S$ scales ($S$ equals to 6, 5, and 3 for the first, second, and third layer respectively) with rescaling factors $2^{-i/2}, i=0,1,...,S$. Specifically, Random\footnotemark[3] feature, RICA \cite{le2011ica} feature and our DDSFL features were extracted with step size 3 for the first layer, and step size 6 for the second and the third layer. SIFT features \cite{lowe2004distinctive} were extracted from 16x16 patches, the step size was 3. HOG2x2 features \cite{xiao2010sun} were extracted based on cells of size 8x8 with stride of 1 cell. LBP features \cite{ojala2002multiresolution} were extracted from cells of size 8x8.

\footnotetext[3]{DDSFL feature extraction scheme without training procedure, the filter bank $W$ for each layer is randomly generated}

\begin{table*}[!t] 
\begin{center}
\begin{tabular}{|l|c|c|c|}
\hline
Methods & Scene 15 & UIUC Sports & MIT Indoor\\
\hline\hline
GIST\cite{oliva2006building}  & 73.28\% & - & 22.00\%\\
CENTRIST \cite{wu2011centrist} & 83.10\% & 78.50\% & 36.90\%\\
SIFT\cite{lowe2004distinctive} & 82.06\% & 85.12\%& 45.86\% \\
HOG2x2\cite{xiao2010sun} & 81.58\% & 83.96\% & 43.76\%\\
LBP\cite{ojala2002multiresolution} & 82.95\% & 80.04\%& 39.25\%\\
OTC\cite{margolinotc} & 84.37\% & - & 47.33\%\\
\hline\hline
Random\footnotemark[3] & 70.91\% & 67.92\% & 21.34\%\\
RICA\cite{le2011ica} & 80.01\% & 82.45\% & 47.76\%\\
DDSFL & 84.42\% & 86.91\% & 52.26\%\\
\hline\hline
Caffe\cite{donahue2013decaf} & 87.99\% & 93.96\% & 58.52\%\\
SIFT+ Caffe & 89.90\% & 95.05\% & 70.51\%\\
DDSFL + Caffe & \textbf{92.81\%} & \textbf{96.78\%} & \textbf{76.23\%}\\
\hline
\end{tabular}
\end{center}
\vspace{-10pt}
\caption{Comparison results between our DDSFL feature and other features. (Caffe is the global feature learned by the deep ConvNets pre-trained on ImageNet.)}
\label{Table-results-feature}
\end{table*}

We randomly picked 4,000 raw pixel value image patches (2,000 for MIT Indoor and PASCAL VOC 2012) from each training image with different scales, and select 10\% of them as training exemplars (refer to Section \ref{Exemplar}). The margin of the hinge loss function in the objective function Equation \ref{full_obj_fun} was fixed as 1. In each DDSFL layer, the parameters $\xi$, $\lambda_1$, $\lambda_2$, $\gamma$ and $\eta$ were initialized as zeros, and they were sequentially learned one after another by applying cross validation (e.g., learn the best $\xi$ first, and then fix the value of $\xi$ and search for the best value of $\lambda_1$). The maximum number of iterations of updating $W$ and $\alpha_c$ in Algorithm \ref{algorithm_DSFL} was set to 5. 

To convert the local features to global image representation, we utilized the LLC framework \cite{wang2010locality} and SPM \cite{lazebnik2006beyond}, the combination of which can generally lead to good classification results. LLC used locality-constrained linear coding to encode local features, and performed max-pooling and linear-SVM afterwards. While SPM utilized rough spatial structure information, and concatenate all the pooled features from different spatial pooling regions. For all the local features, the size of the codebook was fixed as 2,000, and each image was divided into 1x1, 2x2, and 4x4 spatial pooling regions. We have also tested on other coding schemes (e.g. vector quantization) and pooling strategies (e.g. average pooling), our DDSFL consistently outperforms traditional local features.

In all the experiments of DDSFL, if not specified, we report the results of three layer learning framework with NN based exemplar selection method.

\subsection{Comparison with Other Features}
\begin{table*}[!t]\addtolength{\tabcolsep}{-4.3pt}
  \centering
    \begin{tabular}{|lccccccccccc|}
    \hline
     Methods     & aero   & bike   & bird   & boat   & btl    & bus    & car    & cat    & chair  & cow    &  \\
    \hline
    SIFT\cite{lowe2004distinctive}   & 77.8   & 44.4   & 42.4   & 54.0   & 16.0   & 74.7   & 52.1   & 53.0   & 45.6   & 21.4   &  \\
    HOG2x2\cite{xiao2010sun} & 78.3   & 44.8   & 38.9   & 51.6   & 16.6   & 77.0   & 51.2   & 57.3   & 49.2   & 23.5   &  \\
    LBP\cite{ojala2002multiresolution}    & 78.7   & 40.8   & 36.6   & 53.5   & 16.2   & 75.8   & 46.2   & 55.6   & 45.4   & 20.8   &  \\
    Random\footnotemark[3] & 63.0   & 15.4   & 18.8   & 26.7   & 10.4   & 44.6   & 30.2   & 28.7   & 26.3   & \hspace{1ex}9.4    &  \\
    RICA   & 76.1   & 37.2   & 39.1   & 49.6   & 13.8   & 70.5   & 46.2   & 51.2   & 41.4   & 15.6   &  \\
    DDSFL  & 77.7   & 42.5   & 45.4   & 53.3   & 24.0   & 72.2   & 50.6   & 54.2   & 47.4   & 26.0   &  \\
    Caffe\cite{donahue2013decaf}  & 90.7   & 67.9   & 79.9   & 77.0   & 32.7   & 86.0   & 59.7   & 81.6   & 51.4   & 56.1   &  \\
    SIFT+Caffe & 92.3   & 72.3   & 83.0   & \textbf{79.5}   & 38.1   & \textbf{88.5}   & 65.8   & 84.9   & 59.1   & 61.5   &  \\
    DDSFL+Caffe & \textbf{92.7}   & \textbf{75.4}   & \textbf{85.1}   & 79.4   & \textbf{42.3}   & 88.1   & \textbf{68.5}   & \textbf{87.1}   & \textbf{62.9}   & \textbf{65.8}   &  \\
    \hline\hline
     Methods      & table  & dog    & horse  & moto   & pers   & plant  & sheep  & sofa   & train  & tv     & mAP \\
    \hline
    SIFT\cite{lowe2004distinctive}    & 40.2   & 38.4   & 40.7   & 54.2   & 71.3   & 13.1   & 27.4   & 33.0   & 64.5   & 47.6   & 45.6 \\
    HOG2x2\cite{xiao2010sun} & 40.4   & 40.3   & 40.6   & 46.2   & 72.8   & 13.5   & 34.4   & 34.7   & 63.6   & 49.9   & 46.2 \\
    LBP\cite{ojala2002multiresolution}    & 38.4   & 39.6   & 37.4   & 43.4   & 69.8   & 12.8   & 37.8   & 29.3   & 59.6   & 42.1   & 44.0 \\
    Random\footnotemark[3] & 15.8   & 22.8   & 13.7   & 22.0   & 50.7   & \hspace{1ex}7.3    & 15.1   & 15.8   & 30.0   & 22.9   & 24.5 \\
    RICA\cite{le2011ica}   & 29.8   & 35.6   & 33.3   & 46.6   & 67.0   & 14.1   & 28.9   & 29.6   & 58.9   & 44.3   & 41.4 \\
    DDSFL  & 36.6   & 42.6   & 38.9   & 50.7   & 70.3   & 23.4   & 36.2   & 36.4   & 61.6   & 49.0   & 46.9 \\
    Caffe\cite{donahue2013decaf}  & 55.2   & 72.0   & 69.5   & 75.1   & 88.0   & 36.3   & 63.2   & 43.0   & 87.7   & 64.6   & 66.9 \\
    SIFT+Caffe & \textbf{61.6}   & 76.0   & 72.7   & \textbf{79.0}   & 90.7   & 41.8   & 69.2   & 52.7   & \textbf{88.9}   & 70.1   & 71.4 \\
    DDSFL+Caffe & 60.6   & \textbf{80.2}   & \textbf{74.0}   & 76.4   & \textbf{92.3}   & \textbf{43.6}   & \textbf{74.3}   & \textbf{53.2}   & 88.4   & \textbf{73.7}   & \textbf{73.2} \\
    \hline
    \end{tabular}%
    
  \caption{Class-wise classification comparison results between DDSFL and other features on PASCAL VOC 2012.\label{Table-results-PASCAL}}
\end{table*}

We firstly compare our DDSFL feature with the popular features which have shown good performance on scene images classification. There are three types of features: 1) Hand-crafted features, which include SIFT\cite{lowe2004distinctive}, HoG2x2\cite{xiao2010sun,dalal2005histograms,felzenszwalb2010object}, LBP\cite{ojala2002multiresolution}, GIST\cite{oliva2006building}, and CENTRIST\cite{wu2011centrist}; 2) The baseline features learned in unsupervised way without encoding discriminative or shareable information, which include feature extracted with Random weights\footnotemark[3], and the RICA \cite{le2011ica} feature; 3) State-of-the-art Caffe\cite{donahue2013decaf} feature.

As shown in Table \ref{Table-results-feature} and Table \ref{Table-results-PASCAL}, compared with the hand-crafted features and baseline unsupervised learned features, our DDSFL consistently and significantly outperforms all of them. By comparing with the baseline Random\footnotemark[3] features, the significant performance improvements demonstrate that the learning procedures can help to capture data adaptive information, and help to do better recognition. While by comparing with RICA features, the accuracy increasing indicates the effectiveness of learning discriminative and shareable information from image level labels.

\subsubsection{Complementary Effect with Caffe}

The Caffe feature \cite{donahue2013decaf} shown in Table \ref{Table-results-feature} is a global feature extracted from a deep ConvNets \cite{krizhevsky2012imagenet}, which is pre-trained on ImageNet \cite{deng2010does}. The ConvNets has 7 layers (5 convolution layers + 2 fully connected layers). When applying the pre-trained model to extract features in datasets other than ImageNet, the 6th layer feature generally leads to better results than the 7th layer feature\footnotemark[4] \cite{donahue2013decaf,girshick2013rich}. Therefore, we used the 6th layer Caffe feature for evaluation. Although the Caffe feature is very powerful, it is not fair to directly compare the performance of Caffe and the other features. Firstly, we did not have the huge amount of training data in ImageNet. Secondly, the color information has not been utilized\footnotemark[5]. Thirdly, we focused on learning powerful local features rather than global feature representation, and we supposed these two frameworks should encode complementary information, which was proved to be correct in the last row of Table \ref{Table-results-feature} and Table \ref{Table-results-PASCAL}.

\footnotetext[4]{We also tested with the 7th layer features. We got 87.35\%, 93.44\%, and 58.27\% on the three scene datasets, and got mean accuracy of 65.79\% on PASCAL VOC 2012.}

\footnotetext[5]{We did a small experiment with single layer DDSFL on UIUC sports, the inputs were L*a*b values image patches, around 1\% performance gain can be achieved. However, this performance improvement was achieved with the price of three times of computation power, so we did not test on all the datasets.}

\begin{figure*}[!t]
\begin{center}
\includegraphics[width=0.8\linewidth]{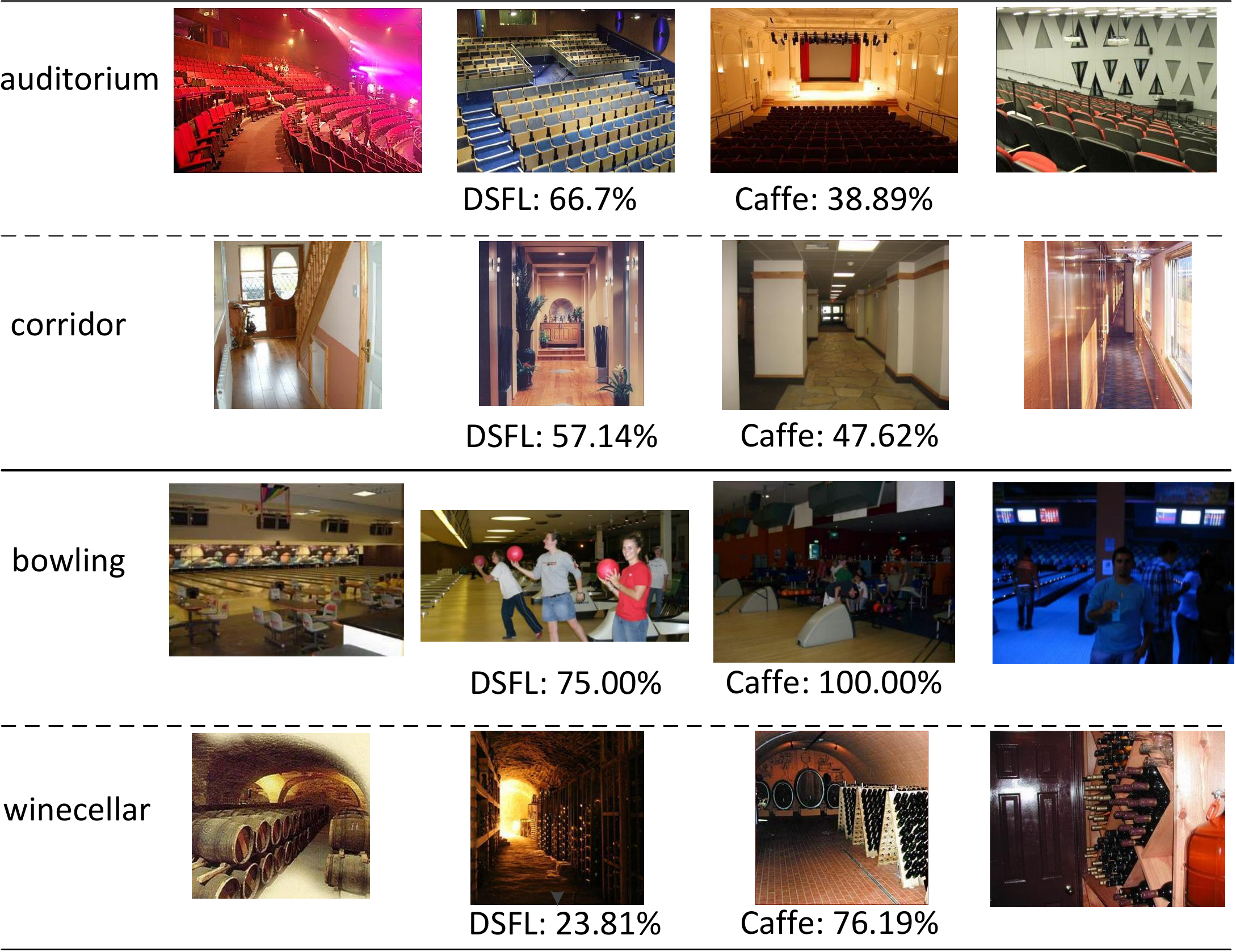}
\end{center}
\vspace{-10pt}
\caption{Comparison results on MIT Indoor. The first two rows show the two categories on which DDSFL works better than Caffe, the last two rows show the classes that are better represented by Caffe. DDSFL and Caffe are complimentary.}
\label{DSFL_vs_DeCAF}
\end{figure*}

To reveal the complementary effect intuitively, we tested on MIT 67 Indoor and compare the class-wise performance. In total, there are 14 categories that DDSFL performed better than Caffe, and 53 categories that Caffe did better than DDSFL. In Figure \ref{DSFL_vs_DeCAF}, the first two rows show the categories that our DDSFL worked better than Caffe, and we show the testing images which were correctly classified by DDSFL, but wrongly classified by Caffe. The last two rows show the categories which Caffe outperformed DDSFL, and we show the testing images which our DDSFL failed to recognize but Caffe could. To quantitatively analyze the complementation effect, we simply concatenate our DDSFL feature (after processing LLC and SPM) with the Caffe feature to check the performance. As shown in the last row of Table \ref{Table-results-feature}, we were able to get a huge performance improvement than purely using the Caffe features and produced the state-of-the-art results. ConvNets apply backpropagation to transmit the supervised information to all the deep layers, this is very helpful for the relatively high layer parameter updating, while it is extremely weak in training bottom layer parameters. In contrast, we explicitly use supervised information to train each layer. Moreover, our method is more suitable for relatively small datasets, as evidenced by the experimental results, while previous attempts on training a CNN classifier directly on small datasets usually failed. Thus, these two works are expected to be complimentary.

To make a fair comparison and exclude the influence brought by coding and pooling schemes, we also tested the combination of SIFT and Caffe. Although this combination also leads to satisfactory results, there is an obvious gap between the performance of SIFT+Caffe and DDSFL+Caffe. This observation indicates that our DDSFL can learn more effective complementary information by encoding data adaptive information. This might because of hand-crafted features such as SIFT usually extracting `garbor-like' features, which are highly similar to the lower level filters learned by ConvNets and not as effectively complementary as our DDSFL.

\begin{table*}[!t] 
\begin{center}
\begin{tabular}{|l|c|c|c|}
\hline
Methods & Scene 15 & UIUC Sports & MIT Indoor\\
\hline\hline
ROI + GIST\cite{quattoni2009recognizing} & - & - & 26.50\% \\
DPM \cite{pandey2011scene} & - & - & 30.40\% \\
Object Bank \cite{li2010object} & 80.90\% & 76.30\% & 37.60\% \\
Discriminative Patches\cite{singh2012unsupervised} & - & - & 38.10\% \\
LDC \cite{wang2013linear} & 80.30\% & - & 43.53\% \\
macrofeatures \cite{boureau2010learning} & 84.30\% & - & -\\
Visual Concepts \cite{liharvesting} & 83.40\% & 84.80\% & 46.40\% \\
SR-LSR \cite{lilatent} & 85.70\% & 83.90\% & -\\
MMDL \cite{wangmax} & 86.35\% & 88.47\% & 50.15\% \\
Discriminative Part Detector\cite{sun2013learning} & 86.00\% & 86.40\% & 51.40\% \\
LScSPM \cite{gao2013laplacian} & 89.78\% & 85.27\% & - \\
IFV \cite{juneja2013blocks} & - & - & 60.77\% \\
MLrep + IFV \cite{doersch2013mid} & - & - & 66.87\% \\
ISPR + IFV \cite{linlearning} & \textbf{91.06\%} & \textbf{92.08\%} & 68.50\% \\
MOP-CNN (Caffe) \cite{gong2014multi} & - & - & \textbf{68.88}\%\\
\hline\hline
DDSFL + Caffe & \textbf{92.81\%} & \textbf{96.78\%} & \textbf{76.23\%}\\
\hline
\end{tabular}
\end{center}
\vspace{-10pt}
\caption{Comparison results of our method and other popular methods on Scene 15, UIUC sports, and MIT Indoor.}
\label{Table-results-sota}
\end{table*}

\subsection{Comparison with Other Methods on Scene Classification}
There are numerous methods applied on the scene image classification, most of them simply utilized the existing hand-crafted features, and focus on the dictionary learning, coding scheme, or mid-level elements mining areas. Most of these works do not conflict with our feature learning framework, our DDSFL feature can easily replace the hand-crafted features, and combine with these methods to achieve better performance. (e.g. LScSPM \cite{gao2013laplacian} and IFV \cite{juneja2013blocks} focused on coding, which can be used to encode our DDSFL features.)

As shown in Table \ref{Table-results-sota}, comparing with the other methods, our DDSFL+Caffe feature achieved the highest accuracy results on all of the three scene classification datasets. Note that Visual Elements \cite{doersch2013mid} utilized numerous patches extracted at scales ranging from 80x80 to the full image size, and the patches were represented by standard HOG \cite{dalal2005histograms} plus a 8x8 color image in L*a*b space, and very high dimensional IFV\cite{juneja2013blocks} features. While MMDL \cite{wangmax} combined 5 types of features on 3 scales. MOP-CNN (Caffe) \cite{gong2014multi}) enhanced the original Caffe with 10\% by considering geometric invariance, which can also be used in our method to further improve the performance.

\subsection{Analysis of The Effects of Different Components}

In this section, we aim to compare our shareable and discriminative learning method to the baseline without encoding such information, which is equivalent to the RICA method in \cite{le2011ica}, and the baseline of using randomly generated feature transformation matrix without learning procedure at all. In the rest of this subsection, we report the results of recruiting different components of the 2 layer DDSFL, which can be efficiently tested and already contain the hierarchical information.

\begin{figure}[!t]
\centering
\subfigure[RICA]{
\label{visualize_rica} 
\includegraphics[width=0.28\linewidth]{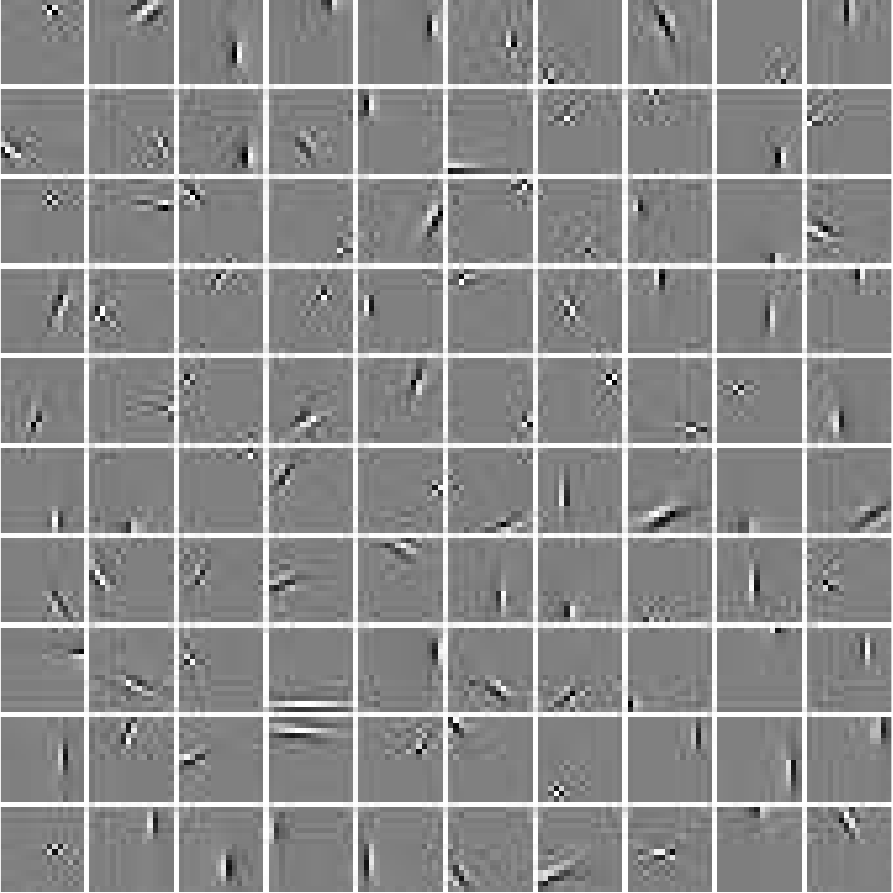}}
\hspace{50pt}
\subfigure[DDSFL]{
\label{visualize_dsfl} 
\includegraphics[width=0.28\linewidth]{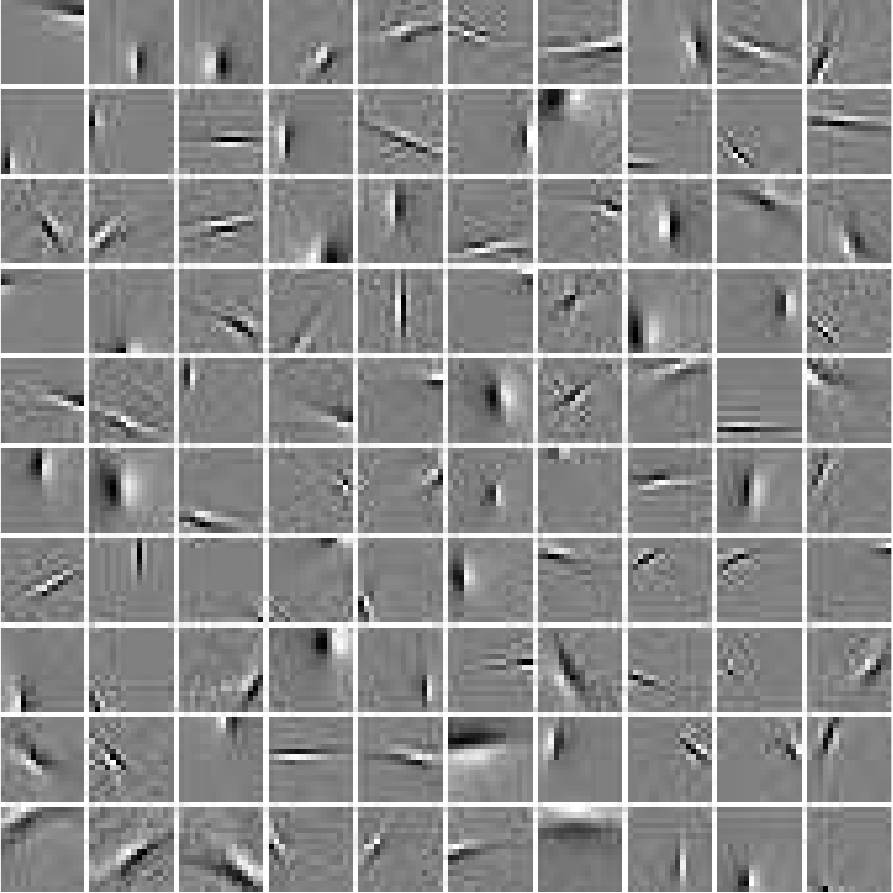}}
\vspace{-5pt}
\caption{Visualization of the filters learned by RICA and DDSFL on the UIUC Sports.} 
\label{visualization} 
\end{figure}

We first show the visualization of the first layer filters learned from UIUC Sports in Figure \ref{visualize_rica} and Figure \ref{visualize_dsfl}. We can see that our DDSFL is able to capture more sharply localized patterns, corresponding to more class-specific visual information. 

\subsubsection{Effect of Learning Shareable Filters}

We tested the DDSFL with or without the feature sharing terms, and got the intermediate results in Table \ref{Table-interations}. The first row of the table shows the result of plainly using randomly generated filters without learning, the poor results indicate that the representation power of the network structure itself is severely limited. The second row shows the unsupervised RICA features learned by solving Equation \ref{psudo_obj_fun_basic}. In the third row, $L_u+L_{\rm{sha}}$ corresponds to features learned with Equation \ref{psudo_obj_fun_sha}. The improvement in accuracy shows that learning shareable features is effective for classification. However, if we removed the global reconstruction error term $\mathcal{L}_u$ and only kept the shareable terms, as shown in the fourth row, the performance dramatically dropped to a random filter bank level.

\subsubsection{Effect of Discriminative Regularization and Exemplar Selection}

Comparing the fifth to seventh rows of Table \ref{Table-interations}, we can observe that without applying exemplars selection step, we could not achieve much improvement because noisy training examples might overwhelm the useful discriminative patterns. However, once using the selected exemplars, our method could achieve significant improvement in classification accuracy. This shows that discriminative exemplar selection is critical in our learning framework. Furthermore, as shown in the last two rows of Table \ref{Table-interations} comparing the NN based and SVM based exemplar selection methods, the former method usually got better performance with higher computational and memory cost. 

Furthermore, it is obvious that only using 10\% of the whole patch set dramatically increased the efficiency of nearest neighbor search in the training step. Thus, our exemplar selection method is both effective and efficient.

\begin{table*}[!t]\addtolength{\tabcolsep}{-2pt}
\begin{center}
\begin{tabular}{|l|c|c|c|}
\hline
Methods & Scene 15 & UIUC Sports & MIT Indoor\\
\hline\hline
Random\footnotemark[3] & 73.52\% & 70.42\% & 23.28\%\\
$L_u$ (RICA\cite{le2011ica}) & 79.85\% & 82.14\% & 47.89\%\\
\hline
$L_u + L_{\rm{sha}}$ & 82.01\% & 83.67\%& 49.70\%\\
$L_{\rm{sha}}$ & 72.69\% & 72.52\%& 24.12\%\\
$L_u + L_{\rm{sha}} + L_{\rm{dis}}$ (without Exemplar) & 82.50\% &  83.43\%& 51.28\%\\
$L_u + L_{\rm{sha}} + L_{\rm{dis}}$ (SVM Exemplar) & 83.02\% & 84.98\% & 51.65\%\\
$L_u + L_{\rm{sha}} + L_{\rm{dis}}$ (NN Exemplar) & \textbf{84.19\%} & \textbf{86.45\%} & \textbf{52.24\%}\\
\hline
\end{tabular}
\end{center}
\vspace{-10pt}
\caption{Analysis of the effect of each component in DDSFL (two layers)}
\label{Table-interations}
\end{table*}

\subsubsection{Effect of Hierarchical Structure}

According to the comparison results shown in Table \ref{Table-hierarchical}, we can observe three interesting phenomena of using different number of layers. Firstly, no matter using how many layers, learning data adaptive features can always achieve more powerful filter banks and lead to better recognition results, and considering shareable and discriminative information can consistently achieve better result than unsupervised learning. Secondly, the good performance of higher layer features heavily depends the high quality lower level features, by using the Random filters, the more layers be processed, the worse results will be achieved. Thirdly, in the RICA and our DDSFL cases, we observe that the performance gain of using two layer structure is much more obvious that than adding the third layer, which might because of lacking of training samples to support lager learning framework (especially when the scale is very small, one image might contain only one 3rd layer patch).

\begin{table*}[!t]
\begin{center}
\begin{tabular}{|l|c|c|c|}
\hline
Methods & Scene 15 & UIUC Sports & MIT Indoor\\
\hline\hline
Random\footnotemark[3] (1 layer) & 74.12\% & 71.46\% & 23.53\%\\
Random\footnotemark[3] (2 layers) & 73.52\% & 70.42\% & 23.28\%\\
Random\footnotemark[3] (3 layers) & 70.91\% & 67.92\% & 21.34\%\\
\hline
RICA\cite{le2011ica} (1 layer) & 77.54\% & 79.03\% & 41.36\%\\
RICA\cite{le2011ica} (2 layers) & 79.85\% & 82.14\% & 47.89\%\\
RICA\cite{le2011ica} (3 layers) & 80.01\% & 82.45\% & 47.76\%\\
\hline
DDSFL (1 layer) & 82.61\% & 83.92\% & 47.16\%\\
DDSFL (2 layers) & 84.19\% & 86.45\% & 52.24\%\\
DDSFL (3 layers) & \textbf{84.42\%} & \textbf{86.91\%} & \textbf{52.26\%}\\
\hline
\end{tabular}
\end{center}
\vspace{-10pt}
\caption{Analysis of the effect of number of layers}
\label{Table-hierarchical}
\end{table*}

\subsubsection{Effect of The Size of Filter Bank}

\begin{figure}[!t]
\begin{center}
\includegraphics[width=0.7\linewidth]{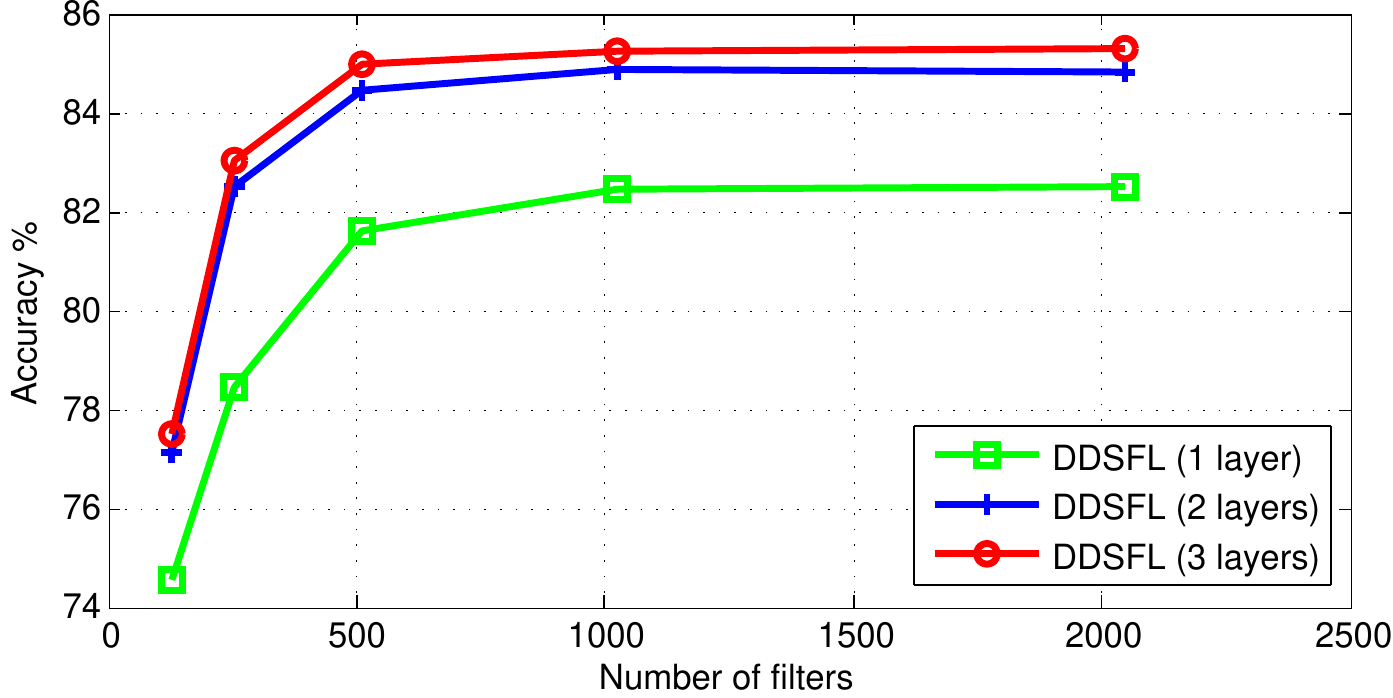}
\end{center}
\vspace{-20pt}
\caption{Results of varying number of filters and layers in Scene 15}
\label{number_filter}
\end{figure}

To further analyze the influence caused by the size of filters, we tested on Scene 15 dataset with 128, 256, 512, 1024, and 2048 filters by using different number of learning layers. The results are shown in Figure \ref{number_filter}. The accuracy results consistently increase when the number of filters increase. When the size of the filter bank is small, especially when the feature transformation matrix is under-complete, the learned features are relatively weak. When the number of filters increases, and $W$ becomes over-complete, the performance is substantially improved. Thus, learning over-complete filter banks do help to obtain better feature representation because the resulting filter banks captures more information. However, when the number of filters further increases, the performance does not increase much, while the learning process will be extremely slow. In our experiment, we learned 400 filters for all the layers as a compromise between efficiency and accuracy. Furthermore, with different number of filters, increasing the number of DDSFL layers also consistently increases the performance, which indicates that higher level DDSFL features can learn complementary patterns. 

\section{Conclusion}
\label{Conclusion}

In this paper, we propose a hierarchical weakly supervised local feature learning method, called DDSFL, to learn discriminative and shareable filter banks to transform local image patches into different visual level features. In our DDSFL method, we learn a flexible number of shared filters to represent shareable patterns exist among similar categories. To enhance the discriminative power, we force the features from the same class to be similar, while features from different classes to be separable. We tested our method on three scene image classification benchmark datasets and PASCAL VOC 2012, the results consistently show that our learned features can outperform most of the existing features. By combining with the Caffe features pre-trained on ImageNet, we can greatly enhance the representation power, and achieve state-of-the-art classification results on all the three scene datasets, and also get promising results on PASCAL.
%
\section*{Acknowledgment}
\label{acknowledgement}
The research is supported by Singapore Ministry of Education (MOE) Tier 2 ARC28/14, and Singapore A*STAR Science and Engineering Research Council PSF1321202099.

\bibliography{bib/egbib}

\end{document}